\documentclass[sigconf, nonacm]{acmart}

\usepackage{amsmath}
\usepackage{array}
\usepackage{colortbl}
\usepackage{tabularx}
\usepackage{enumitem}
\usepackage{longtable}
\usepackage{mdframed}
\usepackage{multirow}
\usepackage{needspace}
\usepackage{placeins}
\usepackage[ruled,vlined]{algorithm2e}
\usepackage{fontawesome5}
\definecolor{linkblue}{RGB}{0,0,238}
\newcommand{\paperlink}[3]{\href{#3}{#1\,\textcolor{linkblue}{#2}}}
\newcommand{\modelurl}{https://huggingface.co/Bobzong/EoG}
\newcommand{\hficon}{\raisebox{-0.15em}{\includegraphics[height=1.0em]{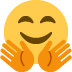}}}

\providecommand{\llbracket}{[\![}
\providecommand{\rrbracket}{]\!]}
\providecommand{\varnothing}{\emptyset}
\definecolor{lightblue}{RGB}{225,240,255}
\definecolor{promptbg}{RGB}{242,244,249}
\definecolor{agentbg}{RGB}{229,245,236}
\definecolor{retrbg}{RGB}{253,236,220}
\definecolor{rlbg}{RGB}{240,231,252}
\definecolor{ablrole}{RGB}{250,234,240}
\definecolor{ablquery}{RGB}{252,246,220}
\definecolor{ablreward}{RGB}{253,242,230}
\definecolor{dswebqsp}{RGB}{250,234,240}
\definecolor{dscwq}{RGB}{252,246,220}
\definecolor{dsgrail}{RGB}{232,245,232}

\definecolor{icwebqsp}{RGB}{196,72,112}
\definecolor{iccwq}{RGB}{196,136,0}
\definecolor{icgrail}{RGB}{52,138,72}
\definecolor{icgtsqa}{RGB}{88,92,190}
\newcommand{\dsicon}[2]{{\setlength{\fboxsep}{0.9pt}\colorbox{#1}{\textcolor{white}{\sffamily\bfseries\tiny #2}}}}
\newcommand{\icW}{\dsicon{icwebqsp}{W}}
\newcommand{\icC}{\dsicon{iccwq}{C}}
\newcommand{\icG}{\dsicon{icgrail}{G}}
\newcommand{\icT}{\dsicon{icgtsqa}{T}}
\newcommand{\best}[1]{\textbf{#1}}
\newcommand{\daulg}[1]{\textbf{#1}}
\mdfdefinestyle{subframe}{%
  backgroundcolor=gray!6,
  linecolor=gray!45,
  linewidth=0.5pt,
  roundcorner=2pt,
  innerleftmargin=6pt,
  innerrightmargin=6pt,
  innertopmargin=5pt,
  innerbottommargin=5pt
}
\AtBeginDocument{%
  }

\setcopyright{none}
\begin{document}

\title{Evolve on Graph: Self-Evolving Knowledge Graph Reasoning without Question Annotations}
\author{Zongyuan Sun\texorpdfstring{$^{1,*}$}{}, Shuai Zhao\texorpdfstring{$^{1,\dagger}$}{}, Ke Shu\texorpdfstring{$^{2,*}$}{}, Ziyue Zhu\texorpdfstring{$^{1}$}{}, Shangyang Wu\texorpdfstring{$^{1}$}{}, Yuanlin Lin\texorpdfstring{$^{1}$}{}, Tao Wang\texorpdfstring{$^{1}$}{}, Shiyue Li\texorpdfstring{$^{1}$}{}, Yingjun Du\texorpdfstring{$^{3}$}{}, Anh Tuan Luu\texorpdfstring{$^{4}$}{}, Haoran Luo\texorpdfstring{$^{4,\dagger}$}{}}
\affiliation{%
  \institution{$^{1}$Beijing University of Posts and Telecommunications, Beijing, China\quad $^{2}$University of Helsinki, Helsinki, Finland}
  \institution{$^{3}$University of Amsterdam, Amsterdam, Netherlands\quad $^{4}$Nanyang Technological University, Singapore}
  \institution{$^{*}$Equal contribution.\quad $^{\dagger}$Corresponding authors.}
  \institution{\vspace{3pt}\normalsize\paperlink{\faHome}{Homepage}{https://szy5.github.io/EoG/}\qquad\paperlink{\faGithub}{GitHub}{https://github.com/Szy5/EoG}\qquad\paperlink{\hficon}{Model}{\modelurl}}
  \country{}
}

\renewcommand{\shortauthors}{Sun et al.}

\begin{abstract}
Knowledge graphs (KGs) provide reliable external knowledge for multi-hop factual reasoning with large language models (LLMs), yet most KGs in real-world domains such as finance and industry lack question--answer (QA) annotations. Without such annotations, existing methods struggle to improve LLM graph reasoning: training-free agents cannot improve their exploration through interaction; reinforcement learning (RL) agents depend on QA data annotated by humans or synthesized by closed-source models; and training data is typically built offline, so its topology mix is fixed and cannot follow the agent's ability. We propose \textbf{Evolve on Graph (EoG)}, a framework that lets an LLM self-evolve its graph reasoning without any QA annotations. Through Graph Self-Play, a single LLM alternates between a Proposer that poses questions and a Solver that answers them. To make questions verifiable, Symbolically Verifiable Topology Synthesis (SVTS) samples reasoning topologies on the KG and executes the corresponding first-order logic queries, yielding topology skeletons with correct and complete answer sets; guided by a skeleton, the Proposer explores the graph to pose a multi-hop question. To let task topologies follow the Solver, a Boundary-Aware Graph Curriculum rewards the Proposer by the Solver's score variance, keeping questions at the Solver's capability boundary so that complex topologies enter training as the Solver improves. On six benchmarks over Freebase, Wikidata, and an industrial KG, EoG consistently outperforms prompting-, retrieval-, and agent-based methods, showing that LLM graph reasoning can improve through self-play without QA annotations. Our code is available at \url{https://github.com/Szy5/EoG}.
\begin{figure}[t]
    \centering
    \vspace{-0.5\baselineskip}
    \includegraphics[width=1.0\linewidth]{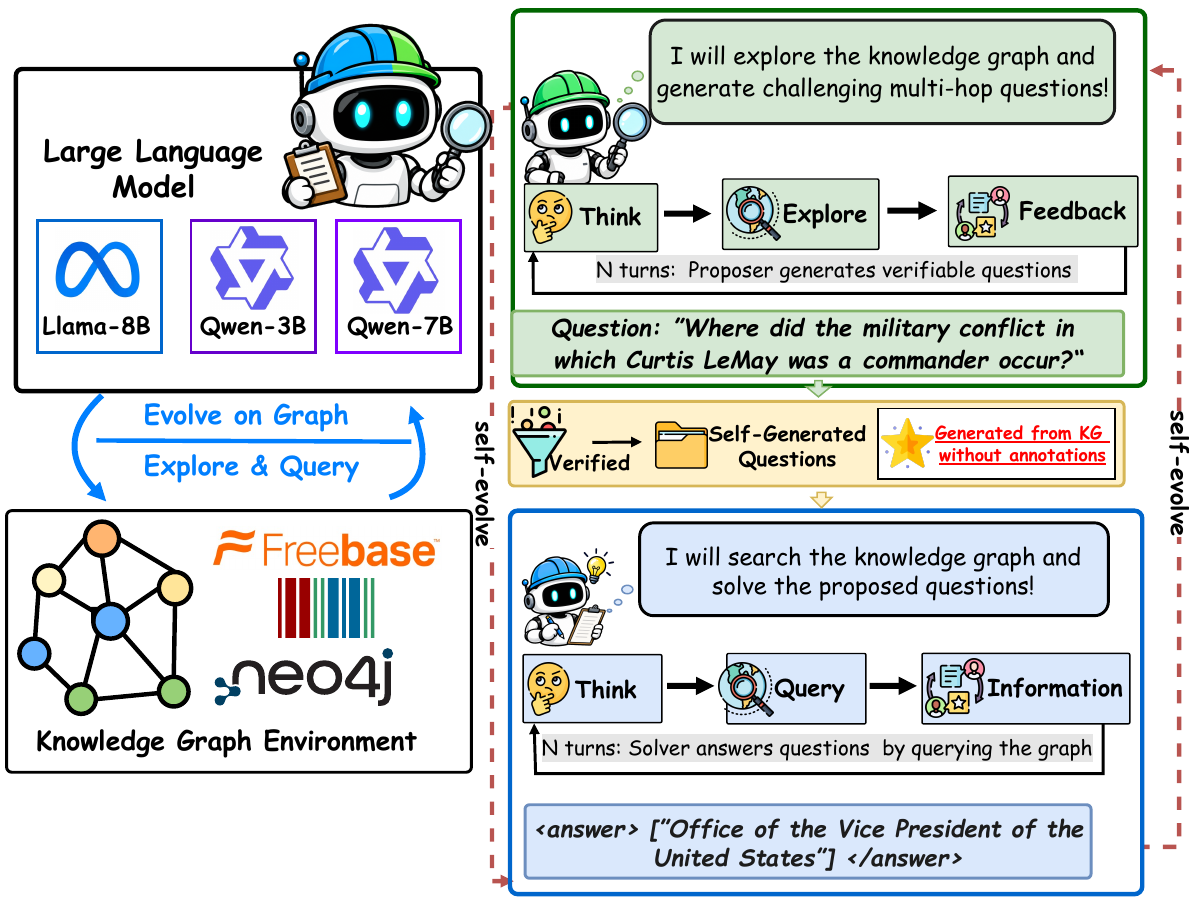}
    \caption{\textbf{With only a KG and no question annotations, EoG lets an LLM self-evolve by proposing questions verified on the graph and then learning to solve them.}}
    \label{fig:overview}
    \vspace{-5mm}
\end{figure}

\end{abstract}




\maketitle

\begin{figure*}[t!]
    \centering
    \includegraphics[width=\textwidth]{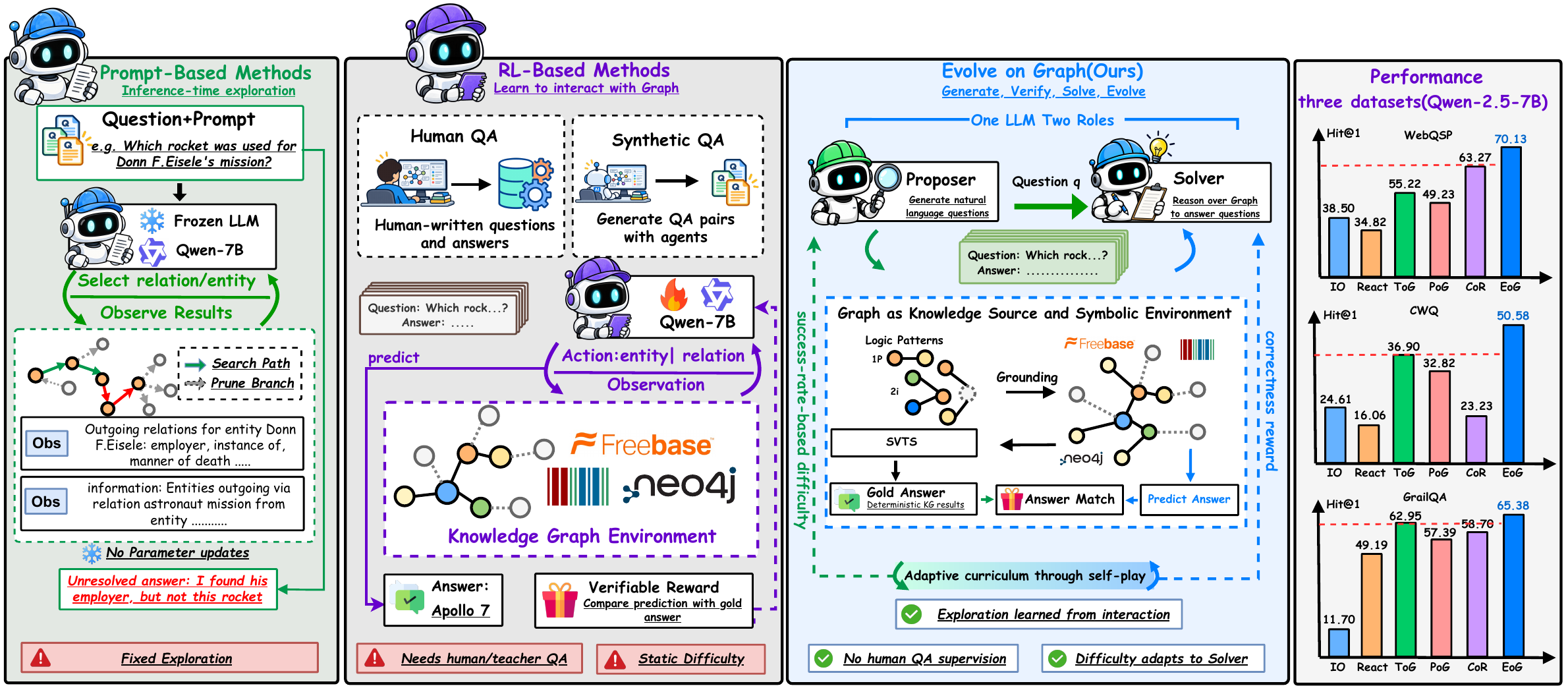}
    \caption{\textbf{Comparison of KG reasoning paradigms.} Prompt-based agents explore with a frozen LLM and RL-based agents learn from fixed QA written by humans or closed-source models, whereas EoG proposes and solves KG-verified tasks of adaptive difficulty. With Qwen-2.5-7B-Instruct, EoG outperforms all shown baselines on three Freebase benchmarks.}
    \label{fig:paradigm}
    \vspace{-5mm}
\end{figure*}
\section{Introduction}
\label{sec:intro}
Large language models (LLMs)~\citep{qwen} have shown remarkable abilities across a wide range of natural language processing tasks~\citep{CoT,Self-Consistency}, yet in complex reasoning that requires deep, multi-hop, and structured factual knowledge, they remain limited by insufficient coverage of long-tail knowledge and by factual hallucinations~\citep{hallucination_1}.

Knowledge graphs (KGs)~\citep{roadmap} store large-scale structured facts and provide reliable external knowledge for LLM reasoning, but this requires the LLM to locate the facts needed at each reasoning step in a large and heterogeneous graph. Research on this problem has moved from one-shot graph retrieval~\citep{RoG,SubGraphRAG,GCR,S-Path-RAG,TARD} toward multi-turn graph exploration (Figure~\ref{fig:paradigm}). Training-free agents~\citep{ToG,PoG,Paths-over-Graph,Open-WorldKG-RAG} prompt a frozen LLM to select relations and entities step by step, while RL-based agents~\citep{KG-R1,GraphWalker,Graph-RFT,ReasoningbyExploration} treat the KG as an interactive environment and train the exploration policy on question--answer data with reinforcement learning (RL). However, most KGs in real-world settings such as industry and healthcare~\citep{KAG,MedRAG} come without any question--answer annotations, and their data often cannot be handed to closed-source models~\citep{PrivGemo}. This raises a question: \emph{can an LLM improve its graph reasoning on a KG using only the KG itself, without any question--answer annotations?}

Answering this question, existing methods still face three challenges. (1) \textbf{The exploration ability of training-free agents cannot improve through interaction.} Training-free agents need no annotations and can be applied to a new KG directly, but their exploration strategy is fixed by the prompt and the frozen backbone, so they cannot accumulate experience from interacting with the graph, and their performance drops sharply on small open-source models~\citep{SLMExploration,RPO-RAG}. (2) \textbf{Training data rely on human annotation or closed-source models.} RL-based agents need large amounts of question--answer data: human annotation~\citep{webqsp,cwq,grailqa,DBLP:conf/nips/ZhangJCCEWWZPWM25,EC-spLLM} requires experts familiar with the KG schema and has to be redone for every new KG, while automatic synthesis~\citep{GraphWalker,GRAIL,GraphScout} relies on closed-source models to synthesize question data. On a KG without annotations that cannot be exposed to closed-source models, neither kind of data is easy to obtain. (3) \textbf{The topology of training tasks cannot follow the agent's ability.} The difficulty of a graph reasoning question lies mainly in its reasoning topology. Existing training data fix the mix of topologies offline before training~\citep{GraphDancer,KnowCoder-A1} and do not adjust it as the agent learns. As the agent grows stronger, simple topologies are already mastered while complex ones remain unsolvable, and neither provides effective learning signals.

To address these challenges, we propose \textbf{Evolve on Graph (EoG)}, a framework in which a single LLM self-evolves on a KG using only the KG itself (Figure~\ref{fig:overview}). First, we propose \textbf{Graph Self-Play}, in which the same LLM alternates between a \emph{Proposer} that poses questions and a \emph{Solver} that answers them. The Proposer explores the graph from real anchor entities, recovers the facts, and writes a question, while the Solver explores the graph to answer it. The two roles share parameters. Second, we propose \textbf{Symbolically Verifiable Topology Synthesis (SVTS)}, which samples reasoning topologies on the graph and executes the corresponding first-order logic queries, giving each task a topology skeleton and a sound and complete answer set. The Proposer explores the graph according to the topology skeleton and writes a multi-hop question. In this way, training tasks and their answers come only from the KG itself, without human annotation or closed-source models. Third, we propose the \textbf{Boundary-Aware Graph Curriculum}, which reinforces the Proposer according to the Solver's score variance, so that it keeps posing questions on the Solver's \emph{capability boundary}, i.e., questions the Solver has not mastered but can still solve. As the Solver grows stronger, more questions on complex topologies fall on its boundary and enter training, forming a graph curriculum that evolves with the Solver.

On six benchmarks over Freebase, Wikidata, and an industrial KG, EoG outperforms prompting-, agent-, and retrieval-based methods without any annotations. With Qwen-2.5-7B-Instruct, EoG exceeds the strongest baseline by +7.46 (Hit@1) and +12.05 (F1) on average, and by +17.80 (Hit@1) on the industrial KG. EoG thus offers a new self-evolving path for training LLM graph reasoning on real-world KGs without any question--answer annotations.

\section{Related Work}
\label{sec:related}
\begin{table*}[t]
    \caption{\textbf{Comparison with representative methods.} Training-free methods keep $\theta_0$, RL agents learn from fixed tasks written by humans or closed-source models, and EoG learns from KG-verified tasks that adapt to the Solver.}
    \centering
    \small
    \setlength{\tabcolsep}{5pt}
    \renewcommand{\arraystretch}{1.0}
    \begin{tabular}{@{}lp{0.29\textwidth}p{0.30\textwidth}p{0.20\textwidth}@{}}
    \toprule
    \textbf{Method} & \textbf{Exploration} (\S\ref{sec:selfplay}) & \textbf{Training Tasks and Answers} (\S\ref{sec:svts}) & \textbf{Task Difficulty} (\S\ref{sec:frontier}) \\
    \midrule
    \rowcolor{agentbg}
    & $a_t\sim\pi_{\theta_0}(\cdot\mid s_t)$ & $\theta=\theta_0$ & \\
    \rowcolor{agentbg}
    \multirow{-2}{*}{ReAct~\citep{React}} & Frozen LLM calls the graph tools. & No training. & \multirow{-2}{*}{No training tasks.} \\
    \rowcolor{agentbg}
    & $\mathcal{B}_{t+1}=\mathrm{top}\text{-}k\;\pi_{\theta_0}(\cdot\mid q,\mathcal{B}_t)$ & $\theta=\theta_0$ & \\
    \rowcolor{agentbg}
    \multirow{-2}{*}{ToG / PoG~\citep{ToG,PoG}} & Prompted beam search or sub-goal planning. & No training. & \multirow{-2}{*}{No training tasks.} \\
    \midrule
    \rowcolor{retrbg}
    & $o=\mathrm{top}\text{-}k\;s\bigl(q,\mathcal{T}_L(\mathcal{E}_Q)\bigr)$ & $\theta=\theta_0$ & \\
    \rowcolor{retrbg}
    \multirow{-2}{*}{Graph Reranker} & Reranks triples near the anchors once. & No training. & \multirow{-2}{*}{No training tasks.} \\
    \midrule
    \rowcolor{rlbg}
    & $a_t\sim\pi_\theta(\cdot\mid s_t)$ & $(q,\mathcal{A})\sim\mathcal{D}_{\mathrm{human}}$ & $d\sim\mathcal{D}$, fixed \\
    \rowcolor{rlbg}
    \multirow{-2}{*}{KG-R1~\citep{KG-R1}} & Multi-turn exploration trained with RL. & Human-annotated QA on one KG. & Given before training. \\
    \rowcolor{rlbg}
    & $\pi_\theta\leftarrow\mathrm{SFT}(\tau_{\mathrm{teacher}})\rightarrow\mathrm{RL}$ & $(q,\mathcal{A})\sim\mathcal{D}_{\mathrm{teacher}}\cup\mathcal{D}_{\mathrm{CWQ}}$ & $d\sim\mathcal{D}$, fixed \\
    \rowcolor{rlbg}
    \multirow{-2}{*}{GraphWalker~\citep{GraphWalker}} & Imitates closed-source trajectories, then RL. & Closed-source walk questions plus CWQ. & Synthesized offline. \\
    \midrule
    \rowcolor{lightblue}
    & $\pi_\theta^{\mathrm{P}},\ \pi_\theta^{\mathrm{S}}$ with shared $\theta$ & $q\sim\pi_\theta^{\mathrm{P}}(\cdot\mid G_Q,\mathcal{E}_Q,\mathcal{G}),\ \mathcal{A}_Q=\llbracket\phi\rrbracket_{\mathcal{G}}$ & $R_{\mathrm{P}}\bigl(\sigma^{2}_{\theta_t}(q)\bigr)$, Gaussian \\
    \rowcolor{lightblue}
    \multirow{-2}{*}{\textbf{EoG (Ours)}} & Both roles explore the KG and learn. & No annotation; answers executed on the KG. & Peaks at the Solver's boundary. \\
    \bottomrule
    \end{tabular}
    \label{tab:comparison}
    \vspace{-2mm}
\end{table*}

\paragraph{Knowledge-graph-augmented reasoning.}
Retrieval-based methods~\citep{RoG,SubGraphRAG,GCR,S-Path-RAG,TARD,RPO-RAG,WhentoTrust,GNN-RAG,PathMind,STEM,DBLP:conf/nips/YuLGTZ25,D-RAG} retrieve paths, triples, or subgraphs in one shot, training-free agents~\citep{ToG,PoG,Paths-over-Graph,Open-WorldKG-RAG,StructGPT,RFKG-CoT,Generate-on-Graph,iQUEST,Enrich-on-Graph,DMAR,CoG-Blueprints,DBLP:conf/acl/PolonuerVANCCZ26,Search-on-Graph,KG-IRAG,GoT} prompt a frozen LLM to explore the graph over multiple turns, and trained agents~\citep{KG-R1,Search-on-Graph-R1,Graph-RFT,ReasoningbyExploration,SymAgent,GraphWalker,KBQA-R1,kg-agent,kbqa-o1,EoG,GraphScout,DBLP:conf/acl/SongSYX26,KG-Reasoner,KG-Hopper} learn exploration from question--answer data with RL. These methods either keep exploration fixed or learn it from annotations on one KG, whereas EoG trains graph reasoning on the target KG alone, without question--answer annotations.

\paragraph{Training data for graph reasoning.}
Benchmarks such as WebQSP~\citep{webqsp}, CWQ~\citep{cwq}, and GrailQA~\citep{grailqa} are annotated by experts familiar with the KG schema at a high cost, and GraphWalker~\citep{GraphWalker} scales up data with random walks and closed-source LLMs; both fix their topology mix before training. Self-play lets LLMs pose and solve their own questions in code and math reasoning~\citep{absolutezero,rzero} and search agents~\citep{searchselfplay,DrZero}, but their knowledge comes from model parameters or unstructured text. Without annotations or closed-source models, EoG generates its own tasks on the KG through self-play, with exact answers from query execution and topologies that advance with the Solver (Table~\ref{tab:comparison}), improving the graph reasoning of LLMs.

\section{Methods}
As shown in Table~\ref{tab:comparison}, existing methods either keep graph exploration fixed, retrieve only once, or train on question--answer data written by humans or closed-source models with fixed difficulty. EoG (Figure~\ref{fig:framework}) consists of Graph Self-Play (Section~\ref{sec:selfplay}), SVTS (Section~\ref{sec:svts}), and the Boundary-Aware Graph Curriculum (Section~\ref{sec:frontier}), which address challenges (1)--(3) of Section~\ref{sec:intro}, respectively.

\begin{figure*}[t!]
    \centering
    \includegraphics[width=\textwidth]{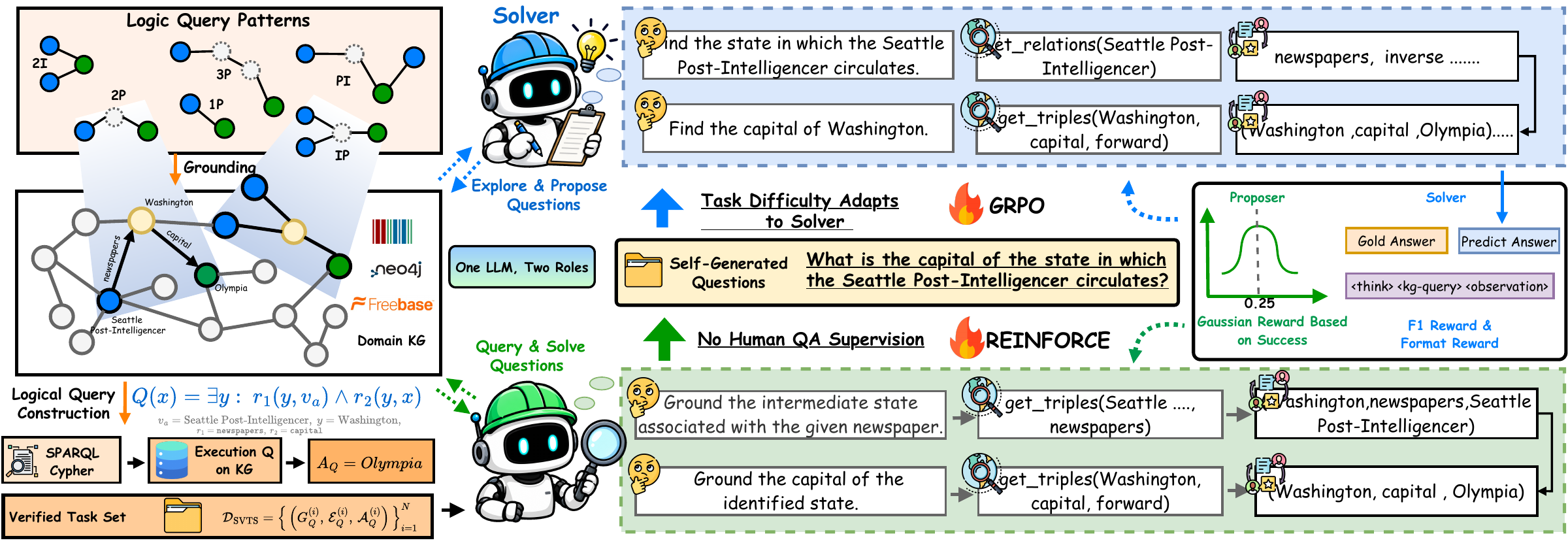}
    \caption{\textbf{Overview of EoG.} One LLM self-plays as the Proposer and the Solver, synthesizes verifiable tasks with SVTS, and forms a boundary-aware curriculum by rewarding the Proposer for questions on the Solver's capability boundary.}
    \label{fig:framework}
\end{figure*}
\subsection{Problem Formulation}
\label{sec:formulation}

\textbf{Definition 1: Logic Query.}
A knowledge graph $\mathcal{G}=(\mathcal{E},\mathcal{R},\mathcal{T})$ has an entity set $\mathcal{E}$, a relation set $\mathcal{R}$, and a triple set $\mathcal{T}\subseteq\mathcal{E}\times\mathcal{R}\times\mathcal{E}$. A multi-hop question corresponds to a first-order logic query
\begin{equation}
\label{eq:query}
\phi(x)=\exists\,y_1,\ldots,y_k:\ e_1\wedge e_2\wedge\cdots\wedge e_n,
\end{equation}
where $x$ is the answer variable, $y_1,\ldots,y_k$ are intermediate variables, and each atom $e_i=r(u,u')$ links an anchor entity or a variable to a variable and is true if and only if $(u,r,u')\in\mathcal{T}$. The answer set of $\phi$ is its denotation $\llbracket\phi\rrbracket_{\mathcal{G}}=\{a\in\mathcal{E}\mid\phi(a)=\mathrm{True}\}$.

\textbf{Definition 2: Graph Environment.}
Following GraphWalker~\citep{GraphWalker}, an agent interacts with $\mathcal{G}$ in a think--query--observe loop through two tools: \texttt{get\_relations} returns the relations of an entity, and \texttt{get\_triples} returns its triples under given relations. At step $t$, given the state $s_t$ of all previous turns, the agent takes an action $a_t$, a tool call or a final answer, and observes triples $o_t\subseteq\mathcal{T}$. A trajectory $\tau=(a_1,o_1,\ldots,a_T)$ collects the evidence
\begin{equation}
\label{eq:evidence}
\mathcal{O}(\tau)=\textstyle\bigcup_{t=1}^{T}o_t\subseteq\mathcal{T},
\end{equation}
where the last action $a_T$ gives the final answer $\mathcal{A}(\tau)\subseteq\mathcal{E}$.

\textbf{Problem Statement.}
Given $\mathcal{G}$ and an initial policy $\pi_{\theta_0}$, we learn a policy $\pi_\theta$ that maximizes the expected F1 of answering multi-hop questions on $\mathcal{G}$, without any question--answer pair or closed-source model. Training uses only symbolic tasks $d=(G_Q,\mathcal{E}_Q,\mathcal{A}_Q)$ built from $\mathcal{G}$, where the query graph $G_Q$ grounds a query $\phi$ on $\mathcal{G}$, $\mathcal{E}_Q$ is its set of anchor entities, and $\mathcal{A}_Q=\llbracket\phi\rrbracket_{\mathcal{G}}$ is its gold answer set.

\subsection{Graph Self-Play}
\label{sec:selfplay}

To let the exploration ability improve through interaction, we propose Graph Self-Play, in which a single policy $\pi_\theta$ plays two roles under different role prompts with shared parameters: the Proposer $\pi_\theta^{\mathrm{P}}$ turns a task into a natural-language question, and the Solver $\pi_\theta^{\mathrm{S}}$ answers it. Both roles reason inside \texttt{<think>}, call the graph tools of Definition~2 inside \texttt{<kg-query>}, and read the returned triples inside \texttt{<information>} blocks, so both act by exploring the same graph. The Solver aims to answer correctly, while the Proposer aims to pose valid questions that the Solver can learn from:
\begin{equation}
\label{eq:game}
\begin{aligned}
&\max_{\pi_\theta^{\mathrm{S}}}\ \mathbb{E}_{d,\;q\sim\pi_\theta^{\mathrm{P}},\;\rho\sim\pi_\theta^{\mathrm{S}}}\bigl[R_{\mathrm{S}}(\rho)\bigr],\\
&\max_{\pi_\theta^{\mathrm{P}}}\ \mathbb{E}_{d,\;q\sim\pi_\theta^{\mathrm{P}}}\bigl[R_{\mathrm{P}}(q)\bigr]\quad\text{s.t.}\quad v(q)=1,
\end{aligned}
\end{equation}
where $d$ is a task synthesized by SVTS, $R_{\mathrm{S}}$ is the Solver reward defined below, $v(q)$ is the validity check of Section~\ref{sec:svts}, and $R_{\mathrm{P}}$ of Section~\ref{sec:frontier} depends on the Solver's performance on $q$. Since the two roles share $\theta$, the policy gains exploration experience both when it poses a question and when it searches for the answers. It includes question generation, question solving, and Solver optimization.

\paragraph{Question generation.}
For each task $d=(G_Q,\mathcal{E}_Q,\mathcal{A}_Q)$, the Proposer reads the reasoning skeleton of $G_Q$, in which only the anchors appear as real entity names, while the intermediate and answer entities appear as placeholders. For example, the skeleton Goodison Park $\xleftarrow{\texttt{arena\_stadium}}$ \texttt{Mid1@} $\xrightarrow{\texttt{owner\_s}}$ \texttt{Ans@} should become ``Who is the owner of the team that plays at Goodison Park?''. Since the tools cannot be called on placeholders, the Proposer explores the graph over multiple turns from the real anchors $\mathcal{E}_Q$ to uncover the facts, and produces a trajectory $\tau^{\mathrm{P}}$ that ends with a question
\begin{equation}
\label{eq:propose}
q\sim\pi_\theta^{\mathrm{P}}(\cdot\mid G_Q,\mathcal{E}_Q,\mathcal{G}),
\end{equation}
where $q$ is extracted from the final \texttt{<result>} block of $\tau^{\mathrm{P}}$. The prompt asks $q$ to name the anchors, keep the full reasoning chain of the skeleton, and hide intermediate and answer entities, so that answering $q$ needs the same multi-hop reasoning as the task.

\paragraph{Question solving.}
For a question $q$, the Solver starts from the anchors $\mathcal{E}_Q$ given in its prompt and samples a group of $n$ trajectories through multi-turn interaction with the graph,
\begin{equation}
\label{eq:solve}
\rho^{j}\sim\pi_\theta^{\mathrm{S}}(\cdot\mid q,\mathcal{E}_Q,\mathcal{G}),\qquad j=1,\ldots,n,
\end{equation}
where the Solver decides which relations and entities to explore and when to stop, and each trajectory ends with a JSON list of entity names in an \texttt{<answer>} block, which gives the prediction $\hat{y}^{j}=\mathcal{A}(\rho^{j})$. Each trajectory is scored by answer and format:
\begin{equation}
\label{eq:rs}
R_{\mathrm{S}}(\rho^{j})=\lambda_{\mathrm{ans}}\,\mathrm{F1}\bigl(\hat{y}^{j},\mathcal{A}_Q\bigr)+\lambda_{\mathrm{fmt}}\,R_{\mathrm{fmt}}(\rho^{j}),
\end{equation}
where the first term is the F1 between the predicted and gold answer sets, $R_{\mathrm{fmt}}\!\in\![0,1]$ measures how well the trajectory follows the think--query--observe structure, and the weights are $\lambda_{\mathrm{ans}}=0.8$ and $\lambda_{\mathrm{fmt}}=0.2$. We use F1 rather than exact match because a task may have several gold answers, so a prediction that recovers only part of $\mathcal{A}_Q$ still receives partial credit. Since $R_{\mathrm{S}}$ scores only the final answer, any exploration path that reaches $\mathcal{A}_Q$ is rewarded, so the Solver learns its own exploration strategy from interaction.

\paragraph{Solver optimization.}
The Solver is updated with GRPO. For a group $\{\rho^{j}\}_{j=1}^{n}$ on the same $q$, the relative advantage is $\hat{A}^{j}=\bigl(R_{\mathrm{S}}(\rho^{j})-\mathrm{mean}\bigr)/\mathrm{std}$, with $\mathrm{mean},\mathrm{std}$ over $\{R_{\mathrm{S}}(\rho^{k})\}_{k=1}^{n}$, and the objective is
\begin{equation}
\label{eq:grpo}
\begin{aligned}
\mathcal{J}_{\mathrm{S}}(\theta)
=\mathbb{E}\!\Bigg[\frac1n\sum_{j=1}^{n}\frac{1}{|\rho^{j}|}
\sum_{t=1}^{|\rho^{j}|}
\min\!\Big(&w^{j}_{t}\hat{A}^{j},
&\operatorname{clip}(w^{j}_{t},1-\epsilon,1+\epsilon)\hat{A}^{j}\Big)\Bigg]\\
-\beta\,\mathbb{D}_{\mathrm{KL}}\!\left[\pi_\theta^{\mathrm{S}}\,\|\,\pi_{\mathrm{ref}}\right],
\end{aligned}
\end{equation}
where $w^{j}_{t}=\pi_\theta^{\mathrm{S}}\bigl(a^{j}_{t}\mid s^{j}_{t}\bigr)/\pi_{\theta_{\mathrm{old}}}^{\mathrm{S}}\bigl(a^{j}_{t}\mid s^{j}_{t}\bigr)$ is the token-level importance ratio at step $t$, $\epsilon$ is the clipping threshold, and $\beta$ is the KL coefficient.

\subsection{Symbolically Verifiable Topology Synthesis}
\label{sec:svts}

To obtain reliable training data without human annotation or closed-source models, we propose SVTS, which views the KG as a first-order logic structure and takes query execution on the graph as the source of gold answers. It includes three steps: defining query patterns, grounding and execution, and question verification.

\paragraph{Query patterns.}
Following complex logical queries over KGs~\citep{ren2020betae}, we define six query patterns as reasoning topologies:
\begin{equation}
\mathcal{P}=\{1p,\,2p,\,3p,\,2i,\,ip,\,pi\},
\end{equation}
where $p$ and $i$ denote projection (one hop along a relation) and intersection (multiple constraints on one variable), respectively; $1p$--$3p$ are $1$- to $3$-hop chain queries, $2i$ is a two-way intersection, and $ip$, $pi$ are nested compositions of projection and intersection. For example, the $2p$ and $2i$ query patterns are written as
\begin{equation}
\label{eq:2p-2i}
\begin{aligned}
\phi_{2p}(x)&=\exists y:\ r_1(v_a,y)\wedge r_2(y,x),\\
\phi_{2i}(x)&=r_1(v_{a_1},x)\wedge r_2(v_{a_2},x),
\end{aligned}
\end{equation}
where $\phi_{2p}$ follows a two-hop chain from one anchor, whereas $\phi_{2i}$ requires the answer to satisfy constraints from two anchors; the logical forms of all six patterns are listed in Appendix~\ref{app:algorithm}. On Freebase, the synthesized questions cover all six patterns, with shares ranging from $9.5\%$ ($pi$) to $22.4\%$ ($1p$), so the Solver must learn path expansion, constraint intersection, and intermediate-variable tracking rather than fit a single pattern (Appendix~\ref{app:contamination}).

\paragraph{Grounding and execution.}
Given a pattern $pat\in\mathcal{P}$, grounding instantiates $pat$ on real facts and extracts its grounded query graphs $\mathcal{I}_{pat}=f_{pat}(\mathcal{G})$, where $f_{pat}$ is the corresponding extraction procedure: it samples a seed entity from a seed pool and walks on $\mathcal{G}$ to bind every variable and relation of $pat$ to a real triple, and it repeats until a target number of tasks per pattern is collected or an attempt budget runs out. The walk only proposes the structure of a task; its answers are decided by execution below. For example, a $2p$ grounded query graph sampled from Freebase is Shaping~$\xrightarrow{\texttt{recipes}}$ Country Sourdough Bread~$\xrightarrow{\texttt{techniques}}$~$x$, whose execution returns two answers, Kneading and Slashing~dough. We then compile each $G_Q\in\mathcal{I}_{pat}$ deterministically into an executable KG query $Q$ for the logic query $\phi$ of Equation~\ref{eq:query}, and evaluate it on $\mathcal{G}$ to obtain the complete gold answer set, i.e., the denotation of $\phi$:
\begin{equation}
\label{eq:exec-main}
\mathcal{A}_Q=\mathrm{Execute}(Q,\mathcal{G})=\llbracket\phi\rrbracket_{\mathcal{G}},
\end{equation}
where queries whose answers are empty, abnormally large, or unnameable are discarded rather than truncated, so that every kept task has an answer set that can be checked objectively. Unlike random-walk synthesis, which takes a single entity of the walk as the answer, execution returns every entity that satisfies the query. This entity is always one of these answers, but when $|\mathcal{A}_Q|>1$, scoring a complete prediction against it alone would give an F1 of only $2/(1+|\mathcal{A}_Q|)$. SVTS runs once per KG before training and only needs query access to the KG, so applying EoG to a new KG requires only its query endpoint. It finally outputs the task set
\begin{equation}
\label{eq:dsvts}
\mathcal{D}_{\mathrm{SVTS}}=\bigl\{(G_Q^{(i)},\mathcal{E}_Q^{(i)},\mathcal{A}_Q^{(i)})\bigr\}_{i=1}^{N},
\end{equation}
where $N$ is the task count (Algorithm~\ref{alg:svts} in Appendix~\ref{app:algorithm}).

\paragraph{Question verification.}
Execution guarantees that $\mathcal{A}_Q$ answers the query $Q$, but not that the question $q$ written by the Proposer expresses the same query. To keep $q$ faithful to its task, we collect the evidence $\mathcal{O}(\tau^{\mathrm{P}})$ retrieved by the Proposer and let the Solver answer $q$ using \emph{only} that evidence as a verifier:
\begin{equation}
\label{eq:verify}
\sigma\sim\pi_\theta^{\mathrm{S}}\bigl(\cdot\mid q,\mathcal{E}_Q,\mathcal{O}(\tau^{\mathrm{P}})\bigr).
\end{equation}
A question is valid only if it passes a rule check $v_{\mathrm{rule}}(q)$, which requires that $q$ can be extracted from the \texttt{<result>} block, has a reasonable length, and names no intermediate or answer entity, and if the verification answer matches the gold answer:
\begin{equation}
\label{eq:valid}
v(q)=v_{\mathrm{rule}}(q)\cdot\mathbb{I}\bigl[\,\mathcal{A}(\sigma)=\mathcal{A}_Q\,\bigr]\in\{0,1\},
\end{equation}
where $\mathbb{I}[\cdot]$ is the indicator function. Only questions with $v(q)=1$ enter reward computation and policy update. The rule check removes questions that leak intermediate or answer entities. The evidence check removes questions that are ambiguous, under-specified, or inconsistent with their task: since the verifier sees only the Proposer's triples, it recovers exactly $\mathcal{A}_Q$ only if these triples, read through $q$, determine the complete answer set. The check is conservative: a faithful question may be rejected when the single verification trajectory $\sigma$ fails, lowering the yield of valid questions, but a question whose evidence does not determine $\mathcal{A}_Q$ is never accepted.

\subsection{Boundary-Aware Graph Curriculum}
\label{sec:frontier}

To let task topologies follow the Solver's ability, we propose the Boundary-Aware Graph Curriculum, which rewards the Proposer by the Solver's score variance, so that it keeps posing questions on the Solver's capability boundary. It includes three parts: the capability boundary, the Proposer reward, and Proposer optimization.

\paragraph{Capability boundary.}
For a valid question $q$ with Solver trajectories $\{\rho^{j}\}_{j=1}^{n}$, let $s_j=\mathrm{F1}(\hat{y}^{j},\mathcal{A}_Q)\in[0,1]$ be the answer score of $\rho^{j}$ in Equation~\ref{eq:rs} and $\bar{s}(q)$ their mean. The Solver's score variance on $q$, which measures how differently its trajectories score, is
\begin{equation}
\label{eq:cbar}
\sigma^{2}(q)=\frac1n\sum_{j=1}^{n}\bigl(s_j-\bar{s}(q)\bigr)^{2}.
\end{equation}
Since the advantages in Equation~\ref{eq:grpo} are normalized within a group, a question on which all $n$ trajectories receive the same score ($\sigma^{2}(q)=0$), such as one the Solver always or never solves, gives almost no learning signal~\citep{DAPO}. The Solver learns mainly from questions with $\sigma^{2}(q)>0$, which it has not mastered but can still partly solve, such as one it answers in full on some trajectories and only in part on others; we call them its \emph{capability boundary}.

\paragraph{Proposer reward.}
We therefore reward the Proposer with a Gaussian boundary reward on the Solver's score variance:
\begin{equation}
\label{eq:rp}
R_{\mathrm{P}}(q)=\exp\!\left(-\frac{\bigl(\sigma^{2}(q)-\tfrac14\bigr)^{2}}{2h}\right),
\end{equation}
where $h>0$ controls the peak width. Since scores lie in $[0,1]$, $\sigma^{2}(q)\le\tfrac14$, with equality only when half of the trajectories score $1$ and the rest $0$. The reward is thus minimal, $\exp(-\tfrac{1}{32h})$, when all scores are equal, higher on the capability boundary, and maximal at such an even split. As the scores are F1 against the complete answer sets of SVTS, a question the Solver answers only in part stays on the boundary instead of counting as solved. The reward also stays low when every trajectory fails, so the Proposer is not reinforced by questions the Solver cannot learn from.

\textbf{Proposition 1.} \textit{With equal format scores, $R_{\mathrm{P}}$ is maximal only where the learning signal peaks and minimal where it vanishes.}\vspace{-2mm}
\begin{proof}
We provide experimental results on the Proposer rewards in Section~\ref{sec:rq-frontier} and a proof in Appendix~\ref{app:proof}.
\end{proof}

\begin{table*}[t!]
    \caption{\textbf{Main results on five public KGQA benchmarks} (Hit@1 and F1, \%; IndustryCarBench in Figure~\ref{fig:carbench}). Full-supervised methods are for reference, with icons marking training sets; bold marks the best annotation-free result; ``--'' means not evaluated.}
    \centering
    \footnotesize
    \setlength{\tabcolsep}{4pt}
    \renewcommand{\arraystretch}{1.05}

    \resizebox{\textwidth}{!}{
    \begin{tabular}{ll*{12}{c}}
        \toprule
        \multirow{2}{*}{\textbf{Method}} & \multirow{2}{*}{\textbf{LLM}}
        & \multicolumn{2}{c}{\textbf{WebQSP}~\icW}
        & \multicolumn{2}{c}{\textbf{CWQ}~\icC}
        & \multicolumn{2}{c}{\textbf{GrailQA}~\icG}
        & \multicolumn{2}{c}{\textbf{GTSQA}~\icT}
        & \multicolumn{2}{c}{\textbf{WikiWebQuestions}}
        & \multicolumn{2}{c}{\textbf{Avg.}} \\
        \cmidrule(lr){3-4} \cmidrule(lr){5-6} \cmidrule(lr){7-8} \cmidrule(lr){9-10} \cmidrule(lr){11-12} \cmidrule(lr){13-14}
        & & \textbf{Hit@1} & \textbf{F1} & \textbf{Hit@1} & \textbf{F1} & \textbf{Hit@1} & \textbf{F1}
        & \textbf{Hit@1} & \textbf{F1} & \textbf{Hit@1} & \textbf{F1} & \textbf{Hit@1} & \textbf{F1} \\
        \midrule
        \rowcolor{gray!15}
        \multicolumn{14}{c}{\textit{Full-Supervised Methods}} \\
        \midrule
        \rowcolor{rlbg}
        KG-R1$^\dagger$~\icW & Qwen-2.5-7B & 78.42 & 69.02 & 50.92 & 46.88 & 61.64 & 59.29 & -- & -- & -- & -- & 63.66 & 58.40 \\
        \rowcolor{rlbg}
        KG-R1$^\dagger$~\icC & Qwen-2.5-7B & 80.71 & 68.57 & 58.03 & 53.93 & 67.81 & 65.52 & -- & -- & -- & -- & 68.85 & 62.67 \\
        \rowcolor{rlbg}
        KG-R1$^\dagger$~\icG & Qwen-2.5-7B & 71.06 & 61.64 & 46.02 & 42.43 & 69.13 & 64.88 & -- & -- & -- & -- & 62.07 & 56.32 \\
        \rowcolor{rlbg}
        KG-R1$^\dagger$~\icT & Qwen-2.5-7B & -- & -- & -- & -- & -- & -- & 48.03 & 49.23 & 75.39 & 66.28 & 61.71 & 57.76 \\
        \midrule
        \rowcolor{gray!15}
        \multicolumn{14}{c}{\textit{Annotation-Free Methods}} \\
        \midrule
        \rowcolor{promptbg}
        & Qwen-2.5-3B & 32.21 & 20.60 & 12.42 & 13.72 & 11.00 & 8.70 & 7.83 & 7.45 & 37.43 & 23.00 & 20.18 & 14.69 \\
        \rowcolor{promptbg}
        & Qwen-2.5-7B & 38.50 & 24.20 & 24.61 & 20.25 & 11.70 & 8.96 & 10.11 & 9.16 & 45.87 & 28.45 & 26.16 & 18.20 \\
        \rowcolor{promptbg} \multirow{-3}{*}{IO} & LLaMA-3.1-8B & 48.14 & 31.79 & 29.85 & 24.30 & 16.30 & 11.13 & 12.95 & 11.56 & 51.99 & 34.09 & 31.85 & 22.57 \\
        \midrule
        \rowcolor{promptbg}
        & Qwen-2.5-3B & 34.96 & 22.20 & 17.76 & 15.00 & 11.00 & 8.70 & 5.73 & 4.92 & 39.83 & 23.41 & 21.86 & 14.85 \\
        \rowcolor{promptbg}
        & Qwen-2.5-7B & 40.94 & 26.98 & 28.38 & 23.70 & 15.40 & 12.25 & 10.23 & 9.15 & 47.30 & 30.27 & 28.45 & 20.47 \\
        \rowcolor{promptbg} \multirow{-3}{*}{CoT~\citep{CoT}} & LLaMA-3.1-8B & 43.99 & 30.01 & 23.93 & 20.18 & 15.20 & 11.16 & 9.06 & 8.27 & 51.86 & 33.87 & 28.81 & 20.70 \\
        \midrule
        \rowcolor{promptbg}
        & Qwen-2.5-3B & 50.64 & 23.50 & 29.31 & 14.99 & 22.10 & 9.81 & 13.13 & 6.40 & 40.50 & 25.47 & 31.14 & 16.03 \\
        \rowcolor{promptbg}
        & Qwen-2.5-7B & 50.58 & 27.94 & 38.97 & 26.08 & 23.00 & 12.67 & 17.76 & 9.90 & 60.46 & 31.19 & 38.15 & 21.56 \\
        \rowcolor{promptbg} \multirow{-3}{*}{SC~\citep{Self-Consistency}} & LLaMA-3.1-8B & 61.81 & 33.77 & 43.39 & 25.78 & 28.00 & 13.80 & 20.84 & 11.85 & 66.71 & 35.70 & 44.15 & 24.18 \\
        \midrule
        \rowcolor{agentbg}
        & Qwen-2.5-3B & 22.63 & 18.25 & 12.06 & 10.63 & 21.26 & 18.61 & 15.47 & 14.72 & 36.46 & 29.56 & 21.58 & 18.35 \\
        \rowcolor{agentbg}
        & Qwen-2.5-7B & 34.82 & 28.42 & 16.06 & 13.45 & 49.19 & 50.07 & 29.41 & 26.52 & 49.14 & 39.63 & 35.72 & 31.62 \\
        \rowcolor{agentbg} \multirow{-3}{*}{ReAct~\citep{React}} & LLaMA-3.1-8B & 48.42 & 40.53 & 17.73 & 15.00 & 27.33 & 28.21 & 18.87 & 17.99 & 33.83 & 28.73 & 29.24 & 26.09 \\
        \midrule
        \rowcolor{agentbg}
        ToG~\citep{ToG} & Qwen-2.5-7B & 55.22 & 43.25 & 36.90 & 35.70 & 62.95 & 56.07 & 38.04 & 34.21 & 64.52 & 49.07 & 51.53 & 43.66 \\
        \rowcolor{agentbg}
        PoG~\citep{PoG} & Qwen-2.5-7B & 49.23 & 36.83 & 32.82 & 28.57 & 57.39 & 50.79 & 35.02 & 32.40 & 53.26 & 40.14 & 45.54 & 37.75 \\
        \rowcolor{agentbg}
        CoR~\citep{CoR} & Qwen-2.5-7B & 63.27 & 57.62 & 23.23 & 21.32 & 58.70 & 54.01 & 26.63 & 24.48 & 66.39 & 54.00 & 47.64 & 42.29 \\
        \rowcolor{retrbg}
        Text-Based Retrieval & Qwen-2.5-7B & 59.86 & 42.78 & 31.61 & 28.78 & 53.64 & 46.60 & 37.73 & 36.44 & 67.22 & 50.00 & 50.01 & 40.92 \\
        \rowcolor{retrbg}
        Graph Reranker & Qwen-2.5-7B & 68.27 & 48.89 & 37.69 & 33.28 & 56.28 & 48.99 & 37.85 & 36.65 & 70.37 & 53.48 & 54.09 & 44.26 \\
        \midrule
        \rowcolor{lightblue}
        & Qwen-2.5-3B & 63.57 & 55.22 & 38.14 & 36.84 & 56.17 & 55.05 & 30.95 & 28.76 & 41.26 & 34.30 & 46.02 & 42.03 \\
        \rowcolor{lightblue} \multirow{-2}{*}{EoG (Qwen-2.5)} & Qwen-2.5-7B & 70.13 & \best{62.10} & \best{50.58} & \best{49.27} & \best{65.38} & \best{63.01} & \best{48.39} & \best{46.36} & \best{73.28} & 60.82 & \best{61.55} & \best{56.31} \\
        \midrule
        \rowcolor{lightblue}
        EoG (LLaMA-3.1) & LLaMA-3.1-8B & \best{71.82} & 61.51 & 50.36 & 48.25 & 63.87 & 62.59 & 36.37 & 34.67 & 70.53 & \best{62.55} & 58.59 & 53.91 \\
        \bottomrule
    \end{tabular}
    }
    \label{tab:main_result}
    \vspace{-2mm}
\end{table*}

\paragraph{Proposer optimization.}
Since the Proposer writes a single question per task, it has no group of samples for a relative advantage; we therefore update it with REINFORCE, taking the Proposer reward as the return of its whole question trajectory:
\begin{equation}
\label{eq:reinforce}
\mathcal{J}_{\mathrm{P}}(\theta)=\mathbb{E}\!\Bigg[
R_{\mathrm{P}}(q)\sum_{t=1}^{|\tau^{\mathrm{P}}|}
\log\pi_\theta^{\mathrm{P}}\!\bigl(a^{\mathrm{P}}_{t}\mid s^{\mathrm{P}}_{t}\bigr)
\Bigg],
\end{equation}
where $a^{\mathrm{P}}_{t}$ and $s^{\mathrm{P}}_{t}$ are the action and state of the Proposer at step $t$. The two roles update alternately: as the Solver masters a kind of question, its scores approach $1$, so $\sigma^{2}$ and $R_{\mathrm{P}}$ fall, driving the Proposer toward harder questions; the boundary thus moves outward and the two roles co-evolve into an adaptive curriculum.

\paragraph{Training loop.}
At each training step, the shared policy first acts as the Proposer on a batch of SVTS tasks and writes one question per task. Each question is checked by Equation~\ref{eq:valid}; questions that fail the check are skipped by both updates, and for every valid question the Solver samples $n$ trajectories on the graph, which yield both the Solver rewards $R_{\mathrm{S}}$ and, through $\sigma^{2}(q)$, the Proposer reward $R_{\mathrm{P}}$. The same parameters $\theta$ are then updated with GRPO on the Solver trajectories and with REINFORCE on the Proposer trajectories. Because both roles are updated at every step, a change in the Solver's ability is reflected in $R_{\mathrm{P}}$ at the next step, so the curriculum follows the Solver without a manually designed difficulty schedule. At test time, only the Solver is used, and it answers human-written questions with the same tools. The whole loop thus uses only the graph: tasks come from SVTS, answers from query execution, and difficulty from the Solver's own score variance on each question.

\section{Experiments}
\label{sec:experiments}

We answer five research questions (RQs). \textbf{RQ1:} Without any annotations, can EoG improve LLM graph reasoning on different KGs? \textbf{RQ2:} Is each component of EoG necessary? \textbf{RQ3:} Can Graph Self-Play improve exploration through interaction? \textbf{RQ4:} Can SVTS replace human annotation as the training data source? \textbf{RQ5:} Can the curriculum advance from simple to complex topologies?

\subsection{Experimental Setup}

\textbf{Datasets.} We evaluate on WebQSP~\citep{webqsp}, CWQ~\citep{cwq}, and GrailQA~\citep{grailqa} over Freebase, GTSQA~\citep{GTSQA} and WikiWebQuestions~\citep{wikiwebquestions} over Wikidata, and IndustryCarBench, built on a real-world automotive-aesthetics KG of car models, design elements, colors, materials, and brands, with 500 expert-written test questions and no training set. Details of all six datasets are in Appendix~\ref{app:datasets}.

\textbf{Baselines.} We compare EoG with prompting-based (IO, CoT~\citep{CoT}, SC~\citep{Self-Consistency}), agent-based (ReAct~\citep{React}, ToG~\citep{ToG}, PoG~\citep{PoG}, CoR~\citep{CoR}), and retrieval-based methods (Text-Based Retrieval, Graph Reranker), all using the same backbone and KG as EoG. GraphWalker~\citep{GraphWalker} synthesizes data with closed-source models, which our setting excludes; the random-walk variant in Table~\ref{tab:ablation} approximates it. KG-R1$^\dagger$ (Section~\ref{sec:rq-svts}) is a fully supervised reference trained on the training set of each benchmark, whose annotations EoG never uses.

\textbf{Implementation Details.} Following prior work~\citep{ToG,KG-R1}, we report Hit@1 and F1. We build EoG on Qwen-2.5-3B/7B-Instruct~\citep{qwen25} and LLaMA-3.1-8B-Instruct~\citep{llama3}, and train it with Verl~\citep{sheng2025hybridflow} for 140 steps on 8 GH200 GPUs. The Solver uses GRPO and the Proposer REINFORCE, with 5 rollouts per task, a learning rate of $1\times10^{-6}$, and at most 15 assistant turns. Unless otherwise stated, analyses use Qwen-2.5-7B-Instruct on the three Freebase benchmarks. Appendix~\ref{app:training} lists the full set of hyperparameters.

\begin{figure*}[t]
\centering
\addtocounter{figure}{1}
\begin{minipage}[t]{0.245\linewidth}
  \centering
  \includegraphics[width=\linewidth]{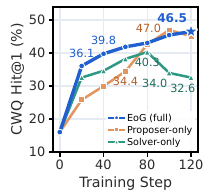}\\[-2pt]
  {\small (a) CWQ Hit@1}
\end{minipage}\hfill
\begin{minipage}[t]{0.245\linewidth}
  \centering
  \includegraphics[width=\linewidth]{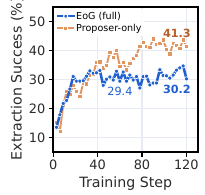}\\[-2pt]
  {\small (b) Extraction rate}
\end{minipage}\hfill
\begin{minipage}[t]{0.245\linewidth}
  \centering
  \includegraphics[width=\linewidth]{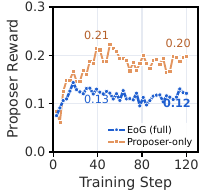}\\[-2pt]
  {\small (c) Proposer reward}
\end{minipage}\hfill
\begin{minipage}[t]{0.245\linewidth}
  \centering
  \includegraphics[width=\linewidth]{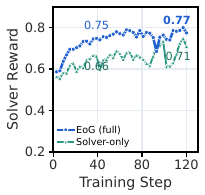}\\[-2pt]
  {\small (d) Solver reward}
\end{minipage}
\caption{\textbf{Training dynamics of role ablations.} Proposer-only and Solver-only denote w/o Solver update and w/o Proposer update. (a) Validation Hit@1 on CWQ. (b) Question-extraction rate of the Proposer. (c--d) Proposer and Solver rewards.}
\label{fig:role_ablation}
\vspace{1mm}
\begin{minipage}[t]{0.48\textwidth}
  \vspace{0pt}
  \centering
  \includegraphics[width=\linewidth]{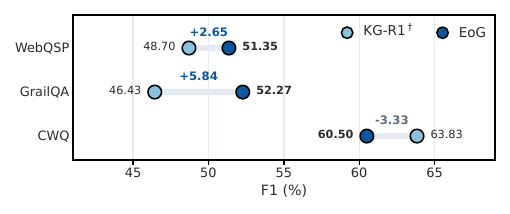}\par
  \refstepcounter{figure}
  \makeatletter
  \def\@currentlabelname{EoG vs. KG-R1 on unseen schema}
  \makeatother
  \label{fig:unseen_schema}
  \addcontentsline{lof}{figure}{\protect\numberline{\thefigure}{\textbf{EoG vs.\ KG-R1$^\dagger$ on unseen schema.} F1 (\%) of EoG and KG-R1$^\dagger$ trained on each dataset with Qwen-2.5-7B-Instruct.}}
  \noindent{\small\raggedright\textbf{Figure \thefigure: EoG vs.\ KG-R1$^\dagger$ on unseen schema. F1 (\%) of EoG and KG-R1$^\dagger$ trained on each dataset with Qwen-2.5-7B-Instruct.}\par}
\end{minipage}
\hfill
\begin{minipage}[t]{0.48\textwidth}
  \vspace{0pt}
  \centering
  \includegraphics[width=\linewidth]{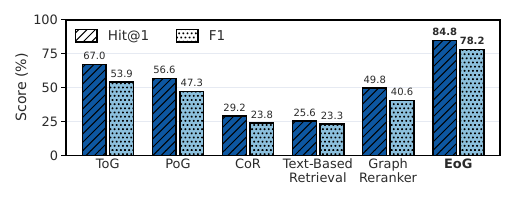}\par
  \addtocounter{figure}{-3}
  \refstepcounter{figure}
  \makeatletter
  \def\@currentlabelname{Results on IndustryCarBench}
  \makeatother
  \label{fig:carbench}
  \addcontentsline{lof}{figure}{\protect\numberline{\thefigure}{\textbf{Results on IndustryCarBench.} Hit@1 and F1 (\%) of EoG and baselines with Qwen-2.5-7B-Instruct.}}
  \noindent{\small\raggedright\textbf{Figure \thefigure: Results on IndustryCarBench. Hit@1 and F1 (\%) of EoG and baselines with Qwen-2.5-7B-Instruct.}\par}
  \addtocounter{figure}{2}
\end{minipage}

\end{figure*}

\subsection{Main Results (RQ1)}
\label{sec:main-results}
As shown in Table~\ref{tab:main_result} and Figure~\ref{fig:carbench}, without any question--answer annotations, EoG enables open-source LLMs to outperform prompting-, agent-, and retrieval-based methods on all three KGs. With Qwen-2.5-7B-Instruct, it exceeds the strongest baseline by \daulg{7.46} points in average Hit@1, and by \daulg{11.61} and \daulg{10.35} points on the multi-hop CWQ and GTSQA, showing that exploration learned through interaction handles complex questions better than retrieval or prompting. On Wikidata, EoG trains directly on the target KG and leads all annotation-free methods in Hit@1 on GTSQA and WikiWebQuestions, so the same pipeline carries over unchanged to a KG with a different schema. On IndustryCarBench, where annotation-based RL agents cannot be trained, EoG trains on the KG alone and exceeds the strongest baseline, ToG, by \daulg{17.80} points in Hit@1, so it applies directly to new KGs that lack annotations. On Qwen-2.5-3B and LLaMA-3.1-8B, against prompting baselines and ReAct, EoG leads in average Hit@1 and F1 (e.g., by 14.44 Hit@1 over SC on LLaMA-3.1-8B), and EoG on Qwen-2.5-3B (46.02) already surpasses PoG on Qwen-2.5-7B (45.54). On schema unseen in the annotations, EoG even outperforms the fully supervised KG-R1$^\dagger$ trained on WebQSP or GrailQA by 2.65 and 5.84 points in F1 (Figure~\ref{fig:unseen_schema}), showing that the exploration it learns through self-play over the whole KG generalizes well beyond the schema covered by annotations.

\subsection{Ablation Study (RQ2)}
\label{sec:ablation}
As shown in Table~\ref{tab:ablation}, we conduct an ablation study on Qwen-2.5-7B-Instruct-based EoG with five ablation settings: removing the Solver update; removing the Proposer update; replacing the query execution of SVTS with random walks~\citep{GraphWalker}, which walk backward from an answer entity and take the start entity as the only answer of the question; and replacing the Gaussian boundary reward with the adversarial reward $1-\bar{c}$ or the triangular boundary reward $1-2|\bar{c}-\frac{1}{2}|$, where $\bar{c}$ is the fraction of Solver trajectories that hit a gold answer. %
All five settings lower the Hit@1 on the multi-hop CWQ by 4.11--14.65 points, showing that Graph Self-Play, SVTS, and the curriculum each matter most on multi-hop questions.

\begin{figure*}[t]
\centering
\begin{minipage}{0.32\linewidth}\centering
  \includegraphics[width=\linewidth]{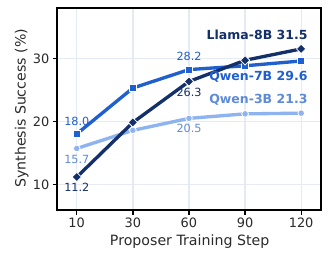}\\{\small (a) Synthesis success}
\end{minipage}\hfill
\begin{minipage}{0.32\linewidth}\centering
  \includegraphics[width=\linewidth]{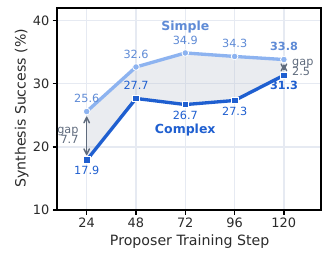}\\{\small (b) Simple vs.\ complex}
\end{minipage}\hfill
\begin{minipage}{0.32\linewidth}\centering
  \includegraphics[width=\linewidth]{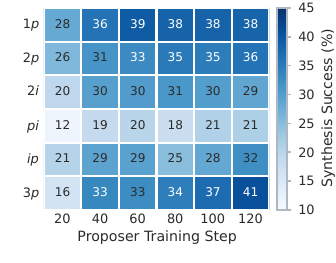}\\{\small (c) Topology heatmap}
\end{minipage}
\caption{\textbf{Evolution of task difficulty.} Synthesis success rate of the Proposer over training steps (a) across three backbones, (b) on simple and complex topologies, and (c) per topology; complex topologies keep improving after simple ones plateau.}
\label{fig:frontier}
\vspace{-3mm}
\end{figure*}
\subsection{Analysis of Graph Self-Play (RQ3)}
\label{sec:rq-selfplay}

\textbf{Comparison with Training-Free Agents.} To examine whether exploration comes from interaction rather than the prompt, we compare EoG with its untrained starting point, ReAct. EoG improves the average Hit@1 of ReAct by more than 24 points on all three backbones and also surpasses ToG (Table~\ref{tab:main_result}), because the exploration strategy of a training-free agent cannot improve from interaction.
\begin{table}[t]
\caption{\textbf{Ablation study of EoG with Qwen-2.5-7B-Instruct on Freebase.} Bold marks the best result in each column.}
\label{tab:ablation}
    \centering
    \footnotesize
    \setlength{\tabcolsep}{2.4pt}
    \renewcommand{\arraystretch}{1.05}

    \begin{tabularx}{\columnwidth}{@{}l*{6}{>{\centering\arraybackslash}X}@{}}
    \toprule
    \multirow{2}{*}{\textbf{Variant}}
    & \multicolumn{2}{c}{\textbf{WebQSP}}
    & \multicolumn{2}{c}{\textbf{CWQ}}
    & \multicolumn{2}{c}{\textbf{GrailQA}}
    \\
    \cmidrule(lr){2-3}
    \cmidrule(lr){4-5}
    \cmidrule(lr){6-7}
    & \textbf{Hit@1} & \textbf{F1}
    & \textbf{Hit@1} & \textbf{F1}
    & \textbf{Hit@1} & \textbf{F1}
    \\
    \midrule
    \rowcolor{agentbg}
    ReAct (untrained)
    & 34.82 & 28.42 & 16.06 & 13.45 & 49.19 & 50.07 \\
    \midrule
    \rowcolor{ablrole}
    w/o Solver update
    & 53.74 & 47.08 & 46.39 & 42.15 & 58.50 & 55.69 \\
    \rowcolor{ablrole}
    w/o Proposer update
    & 63.14 & 56.91 & 36.22 & 35.83 & 55.67 & 55.18 \\
    \rowcolor{ablquery}
    w/o query exec. (Random Walk)
    & 64.87 & 57.67 & 35.93 & 35.64 & 60.12 & 60.41 \\
    \rowcolor{ablreward}
    Adversarial reward
    & 69.51 & 60.40 & 42.59 & 41.85 & 63.26 & 61.01 \\
    \rowcolor{ablreward}
    Triangular reward
    & \best{71.12} & \best{63.73} & 46.47 & 44.03 & 63.46 & 62.10 \\
    \midrule
    \rowcolor{lightblue}
    \textbf{EoG (full, Gaussian reward)}
    & 70.13 & 62.10 & \best{50.58} & \best{49.27} & \best{65.38} & \best{63.01} \\
    \bottomrule
    \end{tabularx}

\end{table}

\textbf{Impact of Removing the Solver Update.} We observe an interesting phenomenon: without Solver updates, the CWQ Hit@1 still rises from 16.06 to 46.39. We believe that proposing a question is itself a multi-hop graph exploration, which transfers to solving via the shared parameters. However, the variant still trails EoG by 16.39 and 6.88 on WebQSP and GrailQA, so the exploration gained from proposing only partly transfers; only when both roles are updated alternately do proposing and solving reinforce each other.

\subsection{Analysis of SVTS (RQ4)}
\label{sec:rq-svts}
To examine whether SVTS can replace human annotation, we design KG-R1$^\dagger$, a variant of KG-R1~\citep{KG-R1} that explores the full Freebase instead of a 2-hop subgraph pre-extracted for each question and differs from EoG only in being trained on annotated questions.

\textbf{Comparison with Human Annotation.} As shown in Table~\ref{tab:main_result}, KG-R1$^\dagger$ is stronger in distribution, but out of distribution, the annotation-free EoG achieves an average F1 3.05 and 3.65 higher than KG-R1$^\dagger$ trained on WebQSP and GrailQA, and trails KG-R1$^\dagger$ trained on 27,623 annotated CWQ questions by only 4.49. To examine generalization beyond the annotated schema, we split the Freebase test sets into seen and unseen schema by whether all relations in their SPARQL query appear in the training set of KG-R1$^\dagger$, and compare them on unseen schema (Figure~\ref{fig:unseen_schema}). EoG exceeds KG-R1$^\dagger$ trained on WebQSP and GrailQA by 2.65 and 5.84 in F1, although both lead on seen schema; KG-R1$^\dagger$ trained on CWQ still leads, but by 3.33 instead of 5.05. We believe that the gains of KG-R1$^\dagger$ from annotation are confined to the schema covered by its training set, whereas EoG trains on SVTS tasks sampled across the whole KG, so its graph reasoning transfers better to unseen schema.

\textbf{Comparison with Random-Walk Synthesis.} SVTS outperforms random walks on all three datasets, by 14.65 in Hit@1 on CWQ (Table~\ref{tab:ablation}), for two reasons. First, random walks favor chain paths and rarely yield the intersection or nested questions that multi-constraint reasoning needs. Second, a random walk takes only its start entity as the answer, although the same relations often reach several entities; correct Solver answers are then judged wrong, so the Proposer can hack the boundary reward with questions whose answers are not unique. SVTS instead executes the query to obtain the complete answer set, which removes such incorrect rewards. Moreover, the synthesized questions rarely resemble the test questions: only 2.28\% of them have a nearest Freebase test question with cosine similarity above 0.90 (Appendix~\ref{app:contamination}), so the gains of EoG do not come from test-set leakage into training.

\subsection{Analysis of the Boundary-Aware Graph Curriculum (RQ5)}
\label{sec:rq-frontier}

\textbf{Impact of Removing the Proposer Update.} Without Proposer updates, task topologies no longer follow the Solver: the Solver reward of this variant keeps rising, yet its CWQ Hit@1 drops after step 80 (Figure~\ref{fig:role_ablation}(a, d)), i.e., the Solver overfits the static task distribution of a Proposer without updates. This indicates that a Proposer without updates cannot shift to topologies that the Solver has not mastered, so the Solver stalls on early, simple questions and trails EoG by 14.36 points in Hit@1 on the multi-hop CWQ.

\textbf{Impact of the Boundary Reward.} We compare three Proposer rewards: the adversarial reward, the triangular boundary reward, and the Gaussian boundary reward. During training, the Solver occasionally produces overlong responses, which are truncated and temporarily lower its reward. With the adversarial reward, the Proposer cannot tell a hard question from such a truncated response and is rewarded most when the Solver performs worst: its reward drop lasts the longest among the three rewards, and its CWQ Hit@1 is 7.99 below that of the Gaussian boundary reward (Table~\ref{tab:ablation}). This shows that the Gaussian boundary reward is not reinforced by the Solver's truncated failures, so the curriculum returns to the Solver's capability boundary sooner after its outputs break down.

\textbf{Evolution of Task Difficulty.} As shown in Figure~\ref{fig:frontier}, the synthesis success rate of the Proposer rises on all three backbones (a), complex topologies rise from 17.9 to 31.3 while simple ones level off near 34, narrowing the gap to 2.5 points (b), and $3p$ keeps rising from 16 to 41 after simple patterns plateau (c), while the Solver reward does not decline (Figure~\ref{fig:role_ablation}(d)). This shows that the Proposer shifts toward complex topologies as the Solver improves, forming an adaptive graph curriculum that advances dynamically from easy to hard questions over time, without a manual schedule.

\subsection{Case Study}
\label{sec:case-study}
Appendix~\ref{app:case-study} shows one complete self-play round on a $2i$ task whose executed answer is Jersey. The Proposer finds that Jèrriais is spoken in Jersey, checks that French is also spoken there, and asks ``In which country are both Jèrriais and French spoken?'', which names both anchors but hides the answer. The Solver retrieves the countries of each language, among them Dominica, Mali, and Chad for French, and keeps only Jersey, the one country shared with Jèrriais. The Solver thus answers by intersecting two entity sets on the graph rather than recalling one fact, the reasoning that $2i$ tasks train.

\section{Conclusion}
\label{sec:conclusion}

We present EoG, in which a single LLM self-evolves its graph reasoning on a KG using only the KG itself, aimed at KGs that have no question--answer annotations and cannot be handed to closed-source models. Graph Self-Play lets one policy explore the graph as both the Proposer and the Solver, SVTS derives verifiable tasks with complete answer sets by executing logic queries on the KG, and the Boundary-Aware Graph Curriculum keeps the questions on the Solver's capability boundary as it shifts from simple to complex topologies. On six benchmarks over three KGs, EoG outperforms prompting-, agent-, and retrieval-based methods, and on schema unseen in the annotations it exceeds KG-R1$^\dagger$ trained on WebQSP or GrailQA. These results show that a KG can supply its own training tasks, answers, and curriculum, offering a direction for self-evolving graph reasoning on real-world KGs that lack annotations, requiring nothing beyond the query endpoint of each new target KG.

\clearpage
\bibliographystyle{ACM-Reference-Format}
\bibliography{references}

\FloatBarrier
\clearpage
\appendix
\newcommand{\casebox}[2]{\par\noindent{\setlength{\fboxsep}{1pt}\colorbox{#1}{\parbox{\dimexpr\linewidth-2\fboxsep\relax}{\raggedright\fontsize{6}{6.5}\selectfont #2}}}\par}
\section{Proof of Proposition 1}
\label{app:proof}
\textbf{Proposition 1.} \textit{With equal format scores, $R_{\mathrm{P}}$ is maximal only where the learning signal peaks and minimal where it vanishes.}\vspace{-2mm}
\begin{proof}
Let $q$ be a valid question with Solver trajectories $\{\rho^{j}\}_{j=1}^{n}$, answer scores $s_j\in[0,1]$ and mean $\bar{s}$ as in Equation~\ref{eq:cbar}, and format scores that all equal $f$. Then $R_{\mathrm{S}}(\rho^{j})=\lambda_{\mathrm{fmt}}f+\lambda_{\mathrm{ans}}s_j$ is a positive affine map of $s_j$, and since the group-normalized advantage is invariant to such maps, $\hat{A}^{j}=(s_j-\bar{s})/\sigma$ with $\sigma=\sqrt{\sigma^{2}(q)}$ when $\sigma>0$, and $\hat{A}^{j}=0$ when $\sigma=0$.

\emph{Learning signal.} We measure the Solver's learning signal on $q$ by the mean absolute advantage of its group,
\begin{equation}
\label{eq:signal}
S(q)=\frac1n\sum_{j=1}^{n}\bigl|\hat{A}^{j}\bigr|.
\end{equation}
If $\sigma^{2}(q)=0$, all scores are equal, every $\hat{A}^{j}=0$, and $S(q)=0$. Otherwise the normalized advantages satisfy $\frac1n\sum_{j}(\hat{A}^{j})^{2}=1$, so by the Cauchy--Schwarz inequality $S(q)\le1$, with equality if and only if all $|\hat{A}^{j}|$ are equal, i.e., half of the scores lie at $\bar{s}+\sigma$ and the other half at $\bar{s}-\sigma$.

\emph{Maximum and minimum.} Since $0\le s_j\le1$, we have $s_j^{2}\le s_j$ and hence $\sigma^{2}(q)=\frac1n\sum_{j}s_j^{2}-\bar{s}^{2}\le\bar{s}(1-\bar{s})\le\tfrac14$, where the first inequality is tight only if every $s_j\in\{0,1\}$ and the second only if $\bar{s}=\tfrac12$. Hence $R_{\mathrm{P}}(q)$ reaches its maximum $1$ only when half of the trajectories score $1$ and the other half score $0$; there $\hat{A}^{j}=\pm1$ and $S(q)=1$, its largest possible value. Conversely, $R_{\mathrm{P}}(q)$ takes its minimum $\exp(-\tfrac{1}{32h})$ exactly when $\sigma^{2}(q)=0$, which is exactly when $S(q)=0$. With $n=5$, an even split is impossible and $\sigma^{2}(q)\le0.24$, attained only by a $2{:}3$ split between scores $0$ and $1$.
\end{proof}

\textbf{Remark.} The adversarial and triangular rewards of Section~\ref{sec:ablation} are functions of the success rate $\bar{c}$. The adversarial reward peaks at $\bar{c}=0$, where every trajectory misses, all scores are $0$, and $S(q)=0$. The triangular reward peaks at $\bar{c}=\tfrac12$ but, like the adversarial reward, does not reflect how much of the answer set each trajectory recovers.

\section{Algorithm Details}
\label{app:algorithm}
\begin{table*}[t]
\centering
\begin{minipage}[t]{0.485\textwidth}
\caption{\textbf{Proposer on a $2i$ task.} It explores both anchors and asks for the country they share.}\label{tab:case-pair-proposer}
\casebox{gray!20}{\textbf{Role:} Proposer \quad \textbf{Task:} Jèrriais $\leftarrow$ location.\allowbreak{}country.\allowbreak{}languages\_\allowbreak{}spoken-- Ans@ ; French --language.\allowbreak{}human\_\allowbreak{}language.\allowbreak{}countries\_\allowbreak{}spoken\_\allowbreak{}in $\rightarrow$ Ans@ \quad \textbf{Executed answer:} Jersey}
\casebox{blue!10}{\texttt{<think>} The Path suggests that Jèrriais is a language spoken in a country, and French is also spoken in that same country. The question should ask for the country where both Jèrriais and French are spoken. Let's explore the knowledge graph to identify the country. \texttt{</think>}}
\casebox{green!10}{\texttt{<kg-query>} get\_\allowbreak{}relations("Jèrriais") \texttt{</kg-query>}}
\casebox{yellow!10}{\texttt{<information>} location.\allowbreak{}country.\allowbreak{}languages\_\allowbreak{}spoken, language.\allowbreak{}human\_\allowbreak{}language.\allowbreak{}main\_\allowbreak{}country, language.\allowbreak{}human\_\allowbreak{}language.\allowbreak{}language\_\allowbreak{}family, language.\allowbreak{}language\_\allowbreak{}family.\allowbreak{}languages, people.\allowbreak{}ethnicity.\allowbreak{}languages\_\allowbreak{}spoken, language.\allowbreak{}human\_\allowbreak{}language.\allowbreak{}countries\_\allowbreak{}spoken\_\allowbreak{}in \texttt{</information>}}
\casebox{blue!10}{\texttt{<think>} The relation 'location.\allowbreak{}country.\allowbreak{}languages\_\allowbreak{}spoken' confirms that Jèrriais is spoken in a country. Let's use 'get\_\allowbreak{}triples' to find out which country it is. \texttt{</think>}}
\casebox{green!10}{\texttt{<kg-query>} get\_\allowbreak{}triples("Jèrriais", ["location.\allowbreak{}country.\allowbreak{}languages\_\allowbreak{}spoken"]) \texttt{</kg-query>}}
\casebox{yellow!10}{\texttt{<information>} [Jersey, location.\allowbreak{}country.\allowbreak{}languages\_\allowbreak{}spoken, Jèrriais] \texttt{</information>}}
\casebox{blue!10}{\texttt{<think>} We have determined that Jèrriais is spoken in Jersey. Now, we need to find out if French is also spoken in Jersey. Let's check the languages spoken in Jersey using the relation 'language.\allowbreak{}human\_\allowbreak{}language.\allowbreak{}countries\_\allowbreak{}spoken\_\allowbreak{}in'. \texttt{</think>}}
\casebox{green!10}{\texttt{<kg-query>} get\_\allowbreak{}triples("Jersey", ["language.\allowbreak{}human\_\allowbreak{}language.\allowbreak{}countries\_\allowbreak{}spoken\_\allowbreak{}in"]) \texttt{</kg-query>}}
\casebox{yellow!10}{\texttt{<information>} [French, language.\allowbreak{}human\_\allowbreak{}language.\allowbreak{}countries\_\allowbreak{}spoken\_\allowbreak{}in, Jersey], [English Language, language.\allowbreak{}human\_\allowbreak{}language.\allowbreak{}countries\_\allowbreak{}spoken\_\allowbreak{}in, Jersey], [Jèrriais, language.\allowbreak{}human\_\allowbreak{}language.\allowbreak{}countries\_\allowbreak{}spoken\_\allowbreak{}in, Jersey] \texttt{</information>}}
\casebox{blue!10}{\texttt{<think>} The information confirms that French is indeed spoken in Jersey. Thus, the answer to the question is Jersey. \texttt{</think>}}
\casebox{lightblue}{\texttt{<result>} \{"question": "In which country are both Jèrriais and French spoken?"\} \texttt{</result>}}
\end{minipage}\hfill
\begin{minipage}[t]{0.485\textwidth}
\caption{\textbf{Solver on the question of Table~\ref{tab:case-pair-proposer}.} It intersects the countries of the two languages.}\label{tab:case-pair-solver}
\casebox{gray!20}{\textbf{Role:} Solver \quad \textbf{Question:} In which country are both Jèrriais and French spoken?}
\casebox{blue!10}{\texttt{<think>} 1. The initial entities are Jèrriais and French. I need to find the country where both Jèrriais and French are spoken. I will start by querying the knowledge graph to find related information about Jèrriais and its spoken locations. \texttt{</think>}}
\casebox{green!10}{\texttt{<kg-query>} get\_\allowbreak{}relations("Jèrriais") \texttt{</kg-query>}}
\casebox{yellow!10}{\texttt{<information>} location.\allowbreak{}country.\allowbreak{}languages\_\allowbreak{}spoken, language.\allowbreak{}human\_\allowbreak{}language.\allowbreak{}main\_\allowbreak{}country, language.\allowbreak{}human\_\allowbreak{}language.\allowbreak{}language\_\allowbreak{}family, language.\allowbreak{}language\_\allowbreak{}family.\allowbreak{}languages, people.\allowbreak{}ethnicity.\allowbreak{}languages\_\allowbreak{}spoken, language.\allowbreak{}human\_\allowbreak{}language.\allowbreak{}countries\_\allowbreak{}spoken\_\allowbreak{}in \texttt{</information>}}
\casebox{blue!10}{\texttt{<think>} I now have some potential relations to explore. I will query the knowledge graph with the first relation "location.\allowbreak{}country.\allowbreak{}languages\_\allowbreak{}spoken" to see if it can provide the country where both languages are spoken. \texttt{</think>}}
\casebox{green!10}{\texttt{<kg-query>} get\_\allowbreak{}triples("Jèrriais", ["location.\allowbreak{}country.\allowbreak{}languages\_\allowbreak{}spoken"]) \texttt{</kg-query>}}
\casebox{yellow!10}{\texttt{<information>} [Jersey, location.\allowbreak{}country.\allowbreak{}languages\_\allowbreak{}spoken, Jèrriais] \texttt{</information>}}
\casebox{blue!10}{\texttt{<think>} The knowledge graph has shown that Jèrriais is spoken in Jersey. Now, to find out if French is also spoken there, I will query the knowledge graph with the relation "language.\allowbreak{}human\_\allowbreak{}language.\allowbreak{}countries\_\allowbreak{}spoken\_\allowbreak{}in" to verify if French is spoken in Jersey. \texttt{</think>}}
\casebox{green!10}{\texttt{<kg-query>} get\_\allowbreak{}triples("French", ["language.\allowbreak{}human\_\allowbreak{}language.\allowbreak{}countries\_\allowbreak{}spoken\_\allowbreak{}in"]) \texttt{</kg-query>}}
\casebox{yellow!10}{\texttt{<information>} [French, language.\allowbreak{}human\_\allowbreak{}language.\allowbreak{}countries\_\allowbreak{}spoken\_\allowbreak{}in, Dominica], [French, language.\allowbreak{}human\_\allowbreak{}language.\allowbreak{}countries\_\allowbreak{}spoken\_\allowbreak{}in, Mali], [French, language.\allowbreak{}human\_\allowbreak{}language.\allowbreak{}countries\_\allowbreak{}spoken\_\allowbreak{}in, Chad], [French, language.\allowbreak{}human\_\allowbreak{}language.\allowbreak{}countries\_\allowbreak{}spoken\_\allowbreak{}in, Jersey], [French, language.\allowbreak{}human\_\allowbreak{}language.\allowbreak{}countries\_\allowbreak{}spoken\_\allowbreak{}in, Ivory Coast] \texttt{</information>}}
\casebox{blue!10}{\texttt{<think>} I have confirmed that French is spoken in Jersey alongside Jèrriais. Therefore, the country where both languages are spoken is Jersey. \texttt{</think>}}
\casebox{lightblue}{\texttt{<answer>} ["Jersey"] \texttt{</answer>}}
\end{minipage}
\end{table*}

\begin{figure}[t]
\centering
\begin{minipage}[t]{\linewidth}
\makeatletter\@twocolumnfalse\makeatother %
\begin{algorithm}[H]
\SetAlgoLined
\DontPrintSemicolon
\KwIn{KG $\mathcal{G}$; pattern set $\mathcal{P}$; seed pool $\mathcal{E}_{\mathrm{seed}}$; per-pattern target $N$; attempt budget $T$; answer cap $\kappa$}
\KwOut{task set $\mathcal{D}_{\mathrm{SVTS}}$}
$\mathcal{D}\leftarrow\varnothing$\;
\ForEach{$pat\in\mathcal{P}$}{
  $n\leftarrow 0$;\quad $t\leftarrow 0$\;
  \While{$n<N$ \textbf{and} $t<T$}{
    $t\leftarrow t+1$;\quad sample seed $s\in\mathcal{E}_{\mathrm{seed}}$\;
    $\pi\leftarrow$ random walk realizing $pat$ from $s$ in $\mathcal{G}$\;
    \lIf{$\pi=\bot$}{\textbf{continue}}
    $G_Q\leftarrow\mathrm{assemble}(pat,\pi)$;\quad $Q\leftarrow\mathrm{Compile}(G_Q)$\;
    $\mathcal{A}_Q\leftarrow\mathrm{Execute}(Q,\mathcal{G})$\;
    \lIf{$\mathcal{A}_Q=\varnothing$ \textbf{or} $|\mathcal{A}_Q|\ge\kappa$ \textbf{or} $\neg\mathrm{Named}(\mathcal{A}_Q)$}{\textbf{continue}}
    $\mathcal{D}\leftarrow\mathcal{D}\cup\{(G_Q,\mathcal{E}_Q,\mathcal{A}_Q)\}$;\quad $n\leftarrow n+1$\;
  }
}
\Return{$\mathcal{D}_{\mathrm{SVTS}}\leftarrow\mathcal{D}$}\;
\caption{Symbolically Verifiable Topology Synthesis}
\label{alg:svts}
\end{algorithm}
\end{minipage}
\vspace{-2mm}
\end{figure}

\paragraph{SVTS}
As shown in Table~\ref{tab:patterns} and Algorithm~\ref{alg:svts}, SVTS repeats grounding, compilation, and execution for each of the six query patterns until it collects $N$ tasks or exhausts $T$ attempts. Grounding walks from a seed entity to bind the variables and relations of the pattern to real facts; the resulting query graph is compiled into an executable query and evaluated on the graph, and queries with empty, at least $\kappa$, or unnameable answers are discarded before training.

\begin{table}[!h]
\caption{\textbf{Basic logic query patterns.} $v_a$: anchor entity; $x$: answer variable; $y_i$: intermediate variables; $r_i$: relations.}
\label{tab:patterns}
\centering
\footnotesize
\setlength{\tabcolsep}{5pt}
\renewcommand{\arraystretch}{1.1}
\newcommand{\patternimg}[1]{%
  \raisebox{-0.5\height}{\includegraphics[scale=0.9]{#1}}%
}

\resizebox{\columnwidth}{!}{%
\begin{tabular}{c c l}
\toprule
\textbf{Pattern} & \textbf{Structure}
& \textbf{Logical query} \\
\midrule

$1p$ &
\patternimg{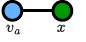} &
$\phi(x)=r_1(v_a,x)$ \\[2pt]

$2p$ &
\patternimg{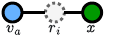} &
$\phi(x)=\exists y:\ r_1(v_a,y)\wedge r_2(y,x)$ \\[2pt]

$3p$ &
\patternimg{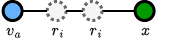} &
$\phi(x)=\exists y_1,y_2:\ r_1(v_a,y_1)\wedge r_2(y_1,y_2)
\wedge r_3(y_2,x)$ \\[2pt]

$2i$ &
\patternimg{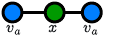} &
$\phi(x)=r_1(v_{a_1},x)\wedge r_2(v_{a_2},x)$ \\[2pt]

$ip$ &
\patternimg{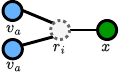} &
$\phi(x)=\exists y:\ r_1(v_{a_1},y)\wedge r_2(v_{a_2},y)
\wedge r_3(y,x)$ \\[2pt]

$pi$ &
\patternimg{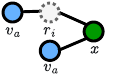} &
$\phi(x)=\exists y:\ r_1(v_{a_1},y)\wedge r_2(y,x)
\wedge r_3(v_{a_2},x)$ \\[2pt]

\bottomrule
\end{tabular}%
}
\end{table}

\section{Experimental Details}
\label{app:experimental-details}
\begin{figure}[t]
\centering
\begin{minipage}{0.32\linewidth}
  \centering
  \includegraphics[width=\linewidth]{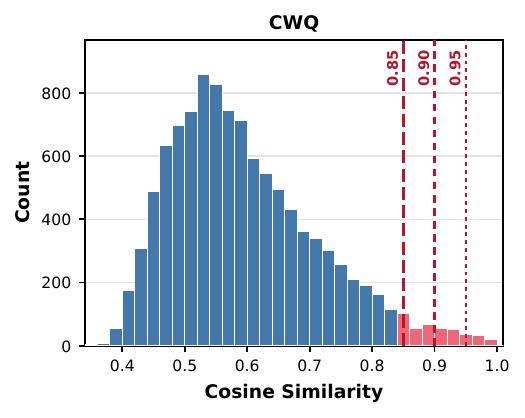}\\[-0.2em]
  {\scriptsize (a) CWQ}
\end{minipage}\hfill
\begin{minipage}{0.32\linewidth}
  \centering
  \includegraphics[width=\linewidth]{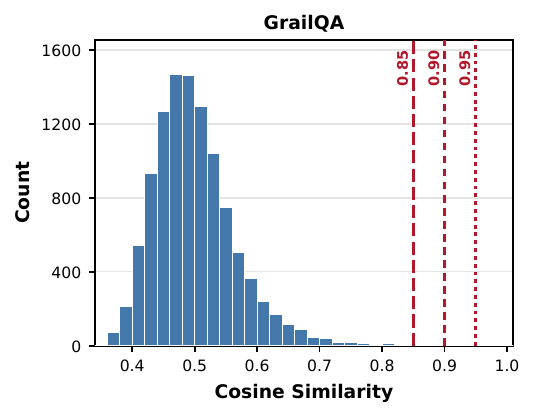}\\[-0.2em]
  {\scriptsize (b) GrailQA}
\end{minipage}\hfill
\begin{minipage}{0.32\linewidth}
  \centering
  \includegraphics[width=\linewidth]{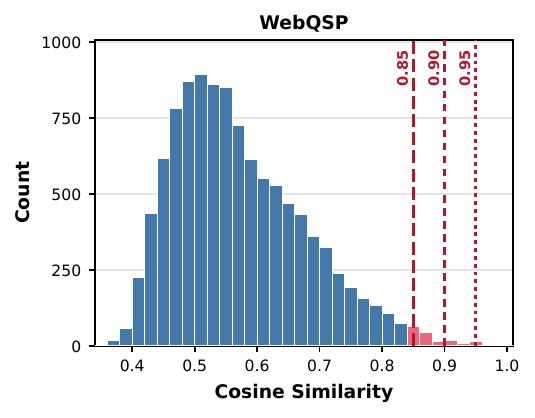}\\[-0.2em]
  {\scriptsize (c) WebQSP}
\end{minipage}
\caption{\textbf{Semantic contamination analysis.} Max cosine similarity of synthesized questions to each test set.}
\label{fig:contamination}
\end{figure}
\begin{table}[t]
\caption{\textbf{Settings of EoG.} Hyperparameters of self-play training with Qwen-2.5-7B-Instruct on Freebase.}
\label{tab:hyperparams}
\centering
\small
\setlength{\tabcolsep}{4pt}
\renewcommand{\arraystretch}{1.0}
\begin{tabular}{lc}
\toprule
\textbf{Parameter} & \textbf{Value} \\
\midrule
Training data & SVTS tasks \\
Training steps & 140 \\
Train / PPO mini-batch size & 256 / 128 \\
Actor learning rate & $1\times10^{-6}$ \\
KL loss coefficient & 0.01 \\
Rollout samples ($n$) & 5 \\
Max response length & 2560 \\
Max assistant turns & 15 \\
Rollout temperature & 0.8 \\
Solver / Proposer estimator & GRPO / REINFORCE \\
Solver reward & $0.8\,\mathrm{F1}+0.2\,R_{\mathrm{fmt}}$ \\
\bottomrule
\end{tabular}
\end{table}

\subsection{Datasets}
\label{app:datasets}
On Freebase, \textbf{WebQSP}~\citep{webqsp} contains single-hop and two-hop questions, \textbf{CWQ}~\citep{cwq} extends WebQuestions with multi-entity constraints, conjunctions, and superlatives, and \textbf{GrailQA}~\citep{grailqa} tests i.i.d., compositional, and zero-shot generalization. On Wikidata, \textbf{GTSQA}~\citep{GTSQA} provides $1{,}622$ test questions with gold answer subgraphs, and \textbf{WikiWebQuestions}~\citep{wikiwebquestions} migrates WebQuestions to Wikidata with re-annotated SPARQL queries. \textbf{IndustryCarBench} is a domain benchmark that we build on a real-world automotive-aesthetics KG covering car models, design elements, attributes, colors, materials, and brands; it contains $500$ test questions written by domain experts, with gold answers executable on the graph and has no annotated training set. EoG never uses the annotated training questions of any benchmark; only KG-R1$^\dagger$ does.

\subsection{Baselines and Metrics}
\label{app:baselines}
\label{app:metrics}
All baselines use the same backbone as the compared EoG model. IO answers directly, CoT~\citep{CoT} reasons step by step, and SC~\citep{Self-Consistency} takes a majority vote over sampled CoT chains. ReAct~\citep{React} interleaves thoughts, graph-tool calls, and observations with the tools and Solver prompt of EoG. ToG~\citep{ToG} runs LLM-guided beam search over entities and relations, PoG~\citep{PoG} decomposes a question into subgoals with adaptive exploration and self-correction, and CoR~\citep{CoR} builds a relation chain and filters the reachable candidates with the question constraints. Text-Based Retrieval selects the top-$k$ verbalized triples by embedding similarity, and Graph Reranker reranks the triples within the $L$-hop neighborhood of the anchors. Following~\citep{ToG,PoG}, Hit@1 checks whether the normalized prediction exactly matches any gold answer, and F1 measures the overlap between the predicted and gold answer lists of each question.

\subsection{Implementation Details}
\label{app:training}

We train EoG with Verl on 8 GH200 GPUs across two nodes, using FSDP with CPU offloading of parameters and optimizer states and SGLang for multi-turn rollouts; Table~\ref{tab:hyperparams} lists the hyperparameters. KG-R1$^\dagger$ converts the official training splits of WebQSP, CWQ, and GrailQA into the Solver format, which yields $3{,}058$, $27{,}623$, and $36{,}954$ questions, and trains one model per dataset. It uses the same backbone, Solver prompt, graph tools, and Freebase endpoint as EoG, so only the source of the training questions differs. Its reward is the F1 of EoG without the format term.

\subsection{Synthetic Data Analysis}
\label{app:contamination}

\paragraph{Topology coverage.}
Synthesized questions on Freebase cover all six patterns: the chain patterns $1p$ ($22.4\%$), $2p$ ($20.6\%$), and $3p$ ($16.5\%$), and the intersection patterns $2i$ ($17.7\%$), $pi$ ($9.5\%$), and $ip$ ($13.3\%$). The Solver must learn path expansion, constraint intersection, and intermediate-variable tracking rather than fit a single pattern.

\paragraph{Domain coverage.}
The questions cover 20 domains, including Places \& Geography ($14.2\%$), Film \& Television ($13.2\%$), Government, Law \& Military ($12.6\%$), People \& Society ($8.7\%$), and Sports ($8.4\%$); the top three together account for only $40.0\%$. Training signals are therefore not confined to a few dense subgraphs of Freebase.

\paragraph{Test-set leakage.}
To rule out test-set leakage, we encode each question with BGE-M3 and take the maximum cosine similarity between each synthetic question and the questions of each Freebase test set (Figure~\ref{fig:contamination}). Only $4.57\%$, $2.28\%$, and $0.71\%$ of the synthetic questions exceed $0.85$, $0.90$, and $0.95$, respectively, and the mean similarity is $0.634$, so there is no meaningful leakage between the synthetic training data and the Freebase evaluation sets.

\section{Case Study}
\label{app:case-study}
Tables~\ref{tab:case-pair-proposer} and~\ref{tab:case-pair-solver} give the full trajectories of the case in Section~\ref{sec:case-study}. The Proposer starts from Jèrriais, finds that it is spoken in Jersey, and queries the languages of Jersey to confirm that French is spoken there before asking. The Solver instead retrieves every country where French is spoken, among them Mali, Chad, and Jersey, and keeps the one shared with Jèrriais, the executed answer.

\end{document}